\documentclass[conference]{IEEEtran}
\IEEEoverridecommandlockouts
\usepackage{cite}
\usepackage{amsmath,amssymb,amsfonts}
\usepackage{algorithmic}
\usepackage{graphicx}
\usepackage{textcomp}
\usepackage{xcolor}
\usepackage{comment}
\usepackage{amsmath}
\usepackage{comment}

\DeclareMathOperator*{\argmin}{arg\,min}

\newcommand{\R}{\mathbb{R}}

\def\bf{\mathbf{f}} 
\newcommand{\be}{\begin{equation}}
\newcommand{\ee}{\end{equation}}                 
\newcommand{\beq}{\begin{eqnarray}}
\newcommand{\eeq}{\end{eqnarray}} 

\def\BibTeX{{\rm B\kern-.05em{\sc i\kern-.025em b}\kern-.08em
    T\kern-.1667em\lower.7ex\hbox{E}\kern-.125emX}}
\begin{document}

\title{Convex Non-negative Matrix Factorization Through Quantum Annealing}

\author{\IEEEauthorblockN{Ahmed ZAIOU}
\IEEEauthorblockA{\textit{EDF Lab Saclay, PERICLES, France }\\
\textit{LIPN - CNRS UMR 7030, Université Sorbonne Paris Nord} \\
\textit{LaMSN, La Maison des Sciences Numériques}\\
\textit{ ahmed.zaiou@edf.fr }\\
}
\and

\IEEEauthorblockN{ Basarab MATEI}
\IEEEauthorblockA{\textit{LIPN - CNRS UMR 7030, Université Sorbonne Paris Nord} \\
\textit{LaMSN, La Maison des Sciences Numériques, France}\\
basarab.matei@sorbonne-paris-nord.fr}

\and

\IEEEauthorblockN{ Younès BENNANI}
\IEEEauthorblockA{\textit{LIPN - CNRS UMR 7030, Université Sorbonne Paris Nord} \\
\textit{LaMSN, La Maison des Sciences Numériques, France}\\
younes.bennani@sorbonne-paris-nord.fr}

\and

\IEEEauthorblockN{ Mohamed HIBTI}
\IEEEauthorblockA{\textit{EDF Lab Saclay, PERICLES} \\
France\\
mohamed.hibti@edf.fr}

} 
\maketitle

\begin{abstract}
In this paper we provide the quantum  version of the Convex Non-negative Matrix Factorization algorithm (Convex-NMF) by using the D-wave quantum annealer. More precisely, we use D-wave 2000Q to find the low rank approximation of a fixed real-valued matrix $X$ by the product of two non-negative matrices factors $W$ and $G$ such that the Frobenius norm of the difference  $X-XWG$ is minimized. In order to solve this optimization problem we proceed in two steps. In the first step we  transform the global real  optimization problem depending on $W,G$ into two quadratic unconstrained binary optimization problems (QUBO) depending on $W $ and $G$ respectively.  In the second  step  we use an  alternative   strategy between the two  QUBO problems corresponding to  $W$ and $G$ to find the global solution. 
The running  of these two QUBO problems on D-wave 2000Q need to use an  embedding to the chimera graph of D-wave 2000Q, this embedding is limited by the number of qubits of D-wave 2000Q. We perform a study on the maximum number of real data to be used by our approach on D-wave 2000Q.  The proposed study is based on the number of qubits used to represent each real variable. We also tested our approach on D-Wave 2000Q with several randomly generated data sets to prove that our approach is faster than the classical approach and also to prove that it gets the best results. 
\end{abstract}


\section{Introduction}
\noindent Data clustering is an unsupervised task whose objective is to determine a finite set of categories (clusters) to define a partition of a data set based on the similarities between its objects. There are several algorithms to perform this task, among them we find the well know K-means, NMF, Semi-NMF and Convex-NMF. All these algorithms are based on some optimization problems, where  some fixed functional  defined by using a fixed metric is minimized. The goal in this strategy is to minimize  the distances between data within the same cluster and maximize the distances between clusters. 
In machine learning  we deal with very high dimensional data and therefore solving optimization problems in high dimension has an huge interest. Optimization in very high dimensions is a true challenge since the near optimal optimization procedures are very slow. In this paper we propose  to use a quantum approach to circumvent this  speed problem related to the high dimension of the data. The quantum approaches allows to 
ameliorate the computational time to find the optimal solution.
In order to define our quantum approach computation, we aim to use   D-wave  quantum annealing to find the global minimum for the functional associated to the Convex-NMF problem. The D-wave  quantum annealing introduced in 1998, allows to solve quadratic unconstrained binary optimization (QUBO) problems. 
We mention that D. Arthur and P. Date in  \cite{arthur2020balanced} show  the equivalence between the balanced K-means on D-wave 2000Q  and some  QUBO problem in a small data set. The authors in  \cite{o2018nonnegative} described the version of non-negative binary matrix factorization algorithm by using  D-Wave quantum annealer.  
The  authors show that  the NMF problem 
$\argmin_{F,G \geq 0} \lVert X - FG \lVert_{F}^{2}$  
where one of the two matrices $F$ and $G$ contains only binary values is equivalent to a QUBO problem.  This last paper was improved to real valued   NMF  by D. Ottaviani and A. Amendola in \cite{ottaviani2018low}, by proposing a method to represent real valued matrix NMF as a QUBO problem. 
\\
\\ In this paper,  we propose a quantum version of the Convex-NMF, to find the global minimum of the functional $\|X - XWG\|_F^{2}$ for a real valued  matrix $X$.
We adopt an {\it alternate optimization} strategy for this functional leading to two independent optimization problems for fixed $G$ and $W $ respectively. 
As already mention, our quantum approach is based on the D-Wave quantum annealer which deal with binary optimization problems. We provide in this paper how to construct the two QUBOs problems to find the two matrices $W$ and $G$.  Both of this QUBOs will be executed in D-wave 2000Q to find the two matrices $W$ and $G$ that minimize the  Frobenius norm $\|X - XWG\|_F^{2}$. This is a constrained optimization problem. The constrains are, on the one hand, on the non-negativity of  all elements of $G$ and $W$ respectively, and  on the  other hand, the sum of  rows/columns of $G/W$ is always equal to $1.$ The rest of this paper is organized as follows: in section \ref{sq:NMF}, we describe the classical and the Convex Non-negative Matrix Factorization principle. 
In section \ref{sq:QA}, we describe the Quantum annealing, the QUBO problem and the binary transformation. 
In section \ref{sq:ourapp}, we present our approach to decompose the convex-NMF on two optimization problems and then we provide the formulation of each one into QUBO. Moreover, we discuss the limitation of the embedding in D-wave 2000Q.
The results obtained after testing our approach are presented in section  \ref{sq:res}. 
Finally the paper ends with a conclusion.

\section{Classical and Convex NMF}
\label{sq:NMF}
\noindent In this section, we shortly describe  the classical version of non-negative matrix factorization and its convex version.
Let $X=(X_{1},X_{2},\ldots,X_{N}) \in
\mathbb{R}^{M \times N}$, be a data matrix with $M$ rows and $N$ columns, here $X_{n} \in \mathbb{R}^{M \times 1}$
represents the $n^{th}$ column of $X$.
In what follows $\|\cdot\|$ stays for the Euclidean norm and $\|\cdot\|_F$ for the Frobenius norm.
Let  $K$  be a fixed input  parameter.
To refer
to the $(m, n)$ element of a matrix $X$, we either write $x_{mn}$ or $X_{mn}$.
Finally, subscripts or summation indices $k$ will be understood to range from $1$
to $K$, subscripts or summation indices $n$ will range from
$1$ up to $N$ (the number of data vectors),
subscripts or summation indices $m$ will range from
$1$ up to $M$ (the dimension of data vectors)
and subscripts or summation primes indices 
will be used to expand inner products between vectors or rows and columns of the same
matrix.

\subsection{Classical NMF}

\noindent In the classical NMF we consider that the data matrix has only  non-negative elements. The classical NMF gives a low rank approximation of   $X$ by the product of two non-negative matrices  
$
  FG
$
where the factors are  
$F = (F_{1},F_{2},\ldots,F_{K}) \in \mathbb{R}_{+}^{M \times K}$, 
$G = (G_{1},G_{2},\ldots,G_{K})^T \in \mathbb{R}_{+}^{K \times N}$ and $T$ denotes the transpose operator.
The NMF decomposition could be formulated as a constrained  optimization problem  by minimizing the following error function:
\begin{equation}
\label{eq:NMF}
 (F,G) = \argmin_{F,G \geq 0}  \|X - FG\|_F^{2}.
\end{equation} 
The non-negativity constraints problem in the matrix form are $F,G \geq 0.$

\subsection{Convex  NMF}
\noindent In the Convex-NMF we consider that the data matrix $X$ is  a real valued matrix. 
 The Convex-NMF problem can be solved if we find the two matrices $W$ and $G$ that minimized the function:
\begin{equation}
\label{eq:CNMF}
     (W,G) = \argmin_{W,G\geq 0} \|X-XWG\|^2_F 
\end{equation}
where $X \in  \R^{M \times N} , W \in  \R_{+}^{N \times K}$ and $ G \in  \R_{+}^{K \times N}$.
In other words, the matrix $W$ represents the positive 
weights coefficients such that we have:
$$F_k=\sum_{n=1}^N w_{nk} X_n =XW_k.$$ 
All the elements in matrices $W$ and $G$ are non-negatives such that:
\begin{equation}
\label{eq:CNMF-c}
\sum_{n=1}^N w_{nk}=1 \mbox{ and } \sum_{n=1}^N g_{kn}=1.
\end{equation}

 \subsection{NMF algorithms}
\noindent The NMF decomposition  has been studied early by Golub and Paatero \cite{paatero1994},\cite{paatero1999}.  Several families of algorithms are proposed to solve this matrix approximation problem. The approach proposed by Lee and Seung \cite{lee2001algorithms} is based on a gradient descent strategy which adaptively define the gradient rates leading to multiplicative update rules. Another solution is to use {\it alternate least square} strategy. We start with a random initialization of $G$ and after that  usually unfolds two phases: at first the functional is minimized with respect to $F$  while in the  second phase, the functional is minimized with respect to $F,$ or $W$ in the convex version.
These  algorithms could converge to a stationary point which are  not necessarily global minima.  There is no guarantee that we can exactly recover the original matrix from $F$ and $G$ or $W$ and $G$ 
so we will approximate it as best as possible in terms of the approximation error measured in Frobenius norm. 
The work of Lee and Seung \cite{lee2001algorithms} revealing that NMF has an inherent clustering property, i.e., it automatically clusters the columns of input matrix $X$
 since the matrix columns vector factor $(F_{1},F_{2},\ldots,F_{K}) $ could be considered as cluster centroids while the
 the matrix rows vector
 factor $(G_{1},G_{2},\ldots,G_{K})^T$ could be considered as cluster indicators.  This fact brought much attention to NMF in machine learning and data
mining communities. 
The aim of clustering is to  cluster
 the columns of $X$, so as to optimize the difference
between $X$ and the clustered matrix revealing
significant block structure.
When $G$ is an orthogonal matrix $G^TG=I$, the  resulting  non-negative
matrix factorization (NMF) is equivalent to relaxed K-means
clustering (see \cite{ding2005equivalence}). 
The $K$-means clustering is one of the most widely used clustering method early developed by Lloyd  \cite{Lloyd82}.
To summarize,  $K$-means clustering problem can be formulated as a matrix approximation
problem \cite{lee2001algorithms}  where the clustering aim is to minimize the approximation
error between the original data $X$ and the reconstructed matrix
based on the cluster structures.

\section{Quantum background}
\label{sq:QA}
\noindent In this section, we introduce the quantum annealing with D-wave 2000Q, we describe  the QUBO problem and we explain the method to present real variables with binary variables.

\subsection{Quantum annealing}
\noindent In order to find the global minimum of an objective function we use the  quantum annealing, which was first proposed by B. Apolloni, N. Cesa Bianchi and D. De Falco in 1988 in the following papers \cite{apolloni1988numerical}\cite{apolloni1989quantum} and has been subsequently reformulated in  \cite{kadowaki1998quantum}\cite{finnila1994quantum}.
 In 2012, D-wave announced the first computer for quantum annealing with $128$ qubits.  This adiabatic quantum computer prepares a Hamiltonian, i.e. prepares a quantum system with several interconnected qubits. These qubits are superposed at the beginning of the processing. The computer will then evolve this Hamiltonian in an adiabatic way to find the solution of the problem. 
Today, it is possible to go up to $2000$ qubits with D-wave 2000Q quantum computer. This computer is deposited by D-wave in open source \cite{finley2017quantum}, it contains the necessary tools for quantum annealing and solves the QUBO problem in a hybrid way on quantum processors and classical hardware architectures using the Qbsolv software.  
D-Wave quantum annealer manipulate the QUBO problems in a native way \cite{mcgeoch2014adiabatic}. It starts with a set of superposed qubits, with each qubit having the same probability of state 0 and state 1. After a few microseconds we found the classical states in the qubits that represent the minimum energy of the problem, or a state very close to it.  
In order to use this computer we just need to transfer the problem to a  QUBO and we do the embedding to give the problem as input to D-wave 2000Q for finding the global minimum.

\subsection{Quadratic Unconstrained Binary Optimization (QUBO)}
\noindent The generic QUBO problem has the following form:

\begin{equation}
    \sum_{b} \psi(b) q_b + \sum_{b<b'} \psi'(b,b') q_b q_{b'}
\end{equation}
where $\psi(b) \in \mathbb{R}$ are the linear coefficients,
$\psi'(b,b') \in \mathbb{R}$ are the quadratic coefficients of the problem and  $q_b, q_{b'} \in \mathbb{B}$  for all $i,j \in [0,n]^2$ where $\mathbb{B}=\{0,1\}$ and  $0\leqslant j \leqslant i \leqslant n $. $n$ is the number of binary variable of the problem. 
\\
\\The problem can be formulated using matrix notation as follows:
\begin{equation}
    \min_{q\in \mathbb{B}^n} q^T\Psi q
\end{equation}
where $ \Psi \in \mathbb{R}^{n\times n}$ is  the symmetric $n\times n$ matrix containing the coefficients $\psi(b)$ and $\psi'(b,b')$ and $q$ it's a binary vector.

 \subsection{Binary representation}
 \label{sq:cQUBO}
\noindent In the QUBO problem, we use only binary variables, so we use the method described by D. Ottaviani and A. Amendola in \cite{ottaviani2018low} to switch from a real representation to a binary representation. 
 In \cite{ottaviani2018low} a generic real element $x_{mn}$ is represented as follows: 
 
 $$x_{mn} =\alpha \sum_{b=0}^B 2^b q_b$$
 with $\alpha$  a constant 
 and $B+1$ is the number of binary variables used to represent the  element $x_{mn}$.
 \\Therefore, to perform a general binary representation of a row of a matrix, we take the matrix $X \in \R^{M\times N}$ where each $X_m \in \R^{1 \times N}$ is the $m$-th row of this matrix. In order to represent this vector with binary variables we represent each item of this vector by $B+1$ binary variables. To do that, we use the following set:
 
 \begin{equation*}
     Q_n \equiv \{n(B+1),n(B+1)+1,\ldots,n(B+1) +B\} 
 \end{equation*}
 
\noindent where $n \in \{0,1,\dots,N\}$. If we take $n=0$, then the first item of the vector $X_m$, is represented by  $Q_0 \equiv \{0,1,\ldots,B\}$ and if we take $n=1$, then the second item of the vector $X_m$, is represented by  $Q_1 \equiv \{(B+1),(B+2),\ldots,2B+1\}$ and so on. So, to represent a row  $X_m$ we reformulate the general transformation as follows:

 \begin{equation}
     X_m = \sum_{n=0}^N \sum_{b=0}^B \beta^m_{nb} q_b
 \end{equation}

 where 
 
\begin{equation*}
\beta^m_{nb}= \left\{ 
\begin{array}{lll}
   \alpha\cdot  2^{b-n(B+1)}  & \text{if} &  b \in Q_n, \\
    0 & \text{if} &  b \notin Q_n.
\end{array}
\right.
\end{equation*}

\noindent Therefore,  we use $(B+1)\times M\times N$ binary variables to represent a matrix $X\in \mathbb{R}^{M\times N}$. 
In the case where $B=9$, $\alpha=0.001$, we can represent every real value  $x_{mn} \in [0,1.023]$.


\section{Our approach}
\label{sq:ourapp}
\noindent In this section we present  our strategy to find the two matrices $W$ and $G$.  
We will use  the {\it alternate optimization} idea to decompose the  initial problem \eqref{eq:CNMF} into two different optimization independent problems.
This strategy gives us the possibility to define two QUBOs problem which  will be solved  separately. Then we show the procedure to get the linear and quadratic coefficients for our problems, at the end we discuss the problem of embedding to the chimera graph of D-wave 2000Q and the maximum size of data that we can use.   
\subsection{Convex-NMF decomposition}
\noindent Firstly, we fix  the matrix 
$W$ and we solve with respect to $G$ the following minimization problem: 
\begin{equation}
\label{eq:CNMFG0}
    \min_{G \geq 0} \|X-XWG\|^2_F 
\end{equation}
secondly, we fix  the matrix $G$ and we solve  with respect to $W$  the following minimization problem: 
\begin{equation}
\label{eq:CNMFW0}
    \min_{W \geq 0} \|X-XWG\|^2_F.  
\end{equation}
In order to satisfy the conditions described in 
\eqref{eq:CNMF-c}, we add two 
constraints on the two problems described by  \eqref{eq:CNMFG0} and  \eqref{eq:CNMFW0} respectively. We rewrite these problems as follows:
\begin{equation}
\label{eq:CNMFG}
    \min_{G} \left(\|X-XWG\|^2_F + \sum_{k=1}^K\left[1- \sum_{n=1}^N g_{kn}\right]^2\right)
\end{equation}
and
\begin{equation}
\label{eq:CNMFW}
    \min_{W} \left(\|X-XWG\|^2_F +\sum_{k=1}^K\left[1- \sum_{n=1}^N\ w_{nk}\right]^2\right)
\end{equation}
Regarding the squared Frobenius norm of a matrix, we recall the following properties:
\begin{subequations}
\label{Frobtrace}
\begin{align}
\left \|X \right\|_F^2&=
\sum_{m=1}^M\sum_{n=1}^N  x_{mn}^2=  \sum_{n=1}^N  \|x_{n}\|^2 \\ & = \sum_{n=1}^N  x^T_{n} x_{n} =\sum_{n=1}^N  (X^T X)_{nn} =Tr(X^TX) 
\end{align}
\end{subequations}
By definition, note that  the functional $\|X-XWG\|^2_F$ take this form:
\begingroup\small
\begin{subequations}
\label{eqF}
\begin{align}
\|X-XWG\|^2_F  & = Tr(X^TX-2G X^TX W +W^TX^T X W G G^T)
\\
& = Tr(X^TX)-2Tr(X^TX W G) \\ &+Tr(W^TX^T X W G G^T)
\end{align}
\end{subequations}
\endgroup
In \eqref{eqF} the first term is constant with respect to 
$G$ and $W$.
\\
\\By direct computations of the second term, we get:
\begingroup\small
\begin{equation}
\label{eqT1}
Tr(X^TX W G) =\sum_{m=1}^M \sum_{n=1}^N \sum_{k=1}^{K} \sum_{n'=1}^N  x_{mn}x_{mn'}w_{n'k} g_{kn}
\end{equation}
\endgroup
The second term is a first order term with respect to $G$ or $W$.
\\
\\For the third term $Tr(W^TX^T X W G G^T)$, we obtain:
\begingroup\small
\begin{equation}
\label{eqT2aa}
  \sum_{k=1}^K \sum_{n=1}^N \sum_{m=1}^M \sum_{n'=1}^N \sum_{k'=1}^K \sum_{n''=1}^N w_{nk}  x_{mn}  x_{mn'}  w_{n'k'} g_{k'n''}  g_{k n''} 
\end{equation}
\endgroup
\subsection{Minimization problem with respect to $G$}
\noindent In this case we fix the matrix $W$ and we are interested to rewrite our functional with respect to the matrix $G$ as variable.
To this end the  \eqref{eqT1} writes:
\begingroup\small
\begin{equation}
\label{eqT1s}
Tr(X^TX W G) =   \sum_{n=1}^N\sum_{k=1}^K (X^TX W)_{nk} g_{kn} 
\end{equation}
\endgroup
while the  \eqref{eqT2aa} writes:
\begin{equation}
\label{eqT2aaa}
Tr(W^TX^T X W G G^T) =\sum_{k=1}^K \sum_{k'=1}^K \sum_{n=1}^N (W^TX^T X W)_{kk'} g_{k'n} g_{kn}
\end{equation}
It follows that:
\begingroup\small
\begin{subequations}
\label{eqp2}
\begin{align}
\|X-XWG\|^2_F &=  Tr(X^TX) -
2\sum_{n=1}^N\sum_{k=1}^K A_{nk} g_{kn}  \\ &+\sum_{k=1}^K \sum_{k'=1}^K \sum_{n=1}^N B_{kk'} g_{k'n} g_{kn}.
\end{align}
\end{subequations}
\endgroup
where $A=X^TX W$ and $B=$ $W^TX^T X W.$ 
\\
\\The constraint $\sum_{k=1}^K \left(1-\sum_{n=1}^N g_{kn}\right)^2$ in  \eqref{eq:CNMFG} can be written as  \eqref{eqp3}.
\begin{equation}
\label{eqp3}
    \sum_{k=1}^K \left( 1  -2   \sum_{n=1}^N g_{kn} +\sum_{n=1}^N   \sum\limits_{\underset{n'\neq n} {n'=1}}^N g_{kn}g_{kn'}  +\sum_{n=1}^N   g^2_{kn} \right)
\end{equation}

\subsection{Minimization problem with respect to $W$}
\noindent In the case of the problem described by 
 \eqref{eq:CNMFW} we fix the matrix $G$ and we are interested to rewrite our functional with the matrix $W$ as variable.
To this end the \eqref{eqT1} writes:
\begin{subequations}
\label{eqT1w}
\begin{align}
    Tr(X^TX W G) &=  Tr(G X^TX W) \\ &=  \sum_{n=1}^N\sum_{k=1}^K (G X^TX)_{kn} w_{nk}
\end{align}
\end{subequations}
\begingroup\small
\begin{subequations}
\label{eqT2awa}
\begin{align}
Tr(W^TX^T X W G G^T) 
 & =\sum_{k=1}^K \sum_{n=1}^N   \sum_{n'=1}^N \sum_{k'=1}^K  D_{nn'}   E_{k' k} w_{nk} w_{n'k'} \\
    &   =\sum_{k=1}^K \sum_{n=1}^N \sum\limits_{\underset{n'\neq n} {n'=1}}^N \sum\limits_{\underset{k'\neq k} {k'=1}}^K  D_{nn'}   E_{k' k} w_{nk} w_{n'k'} \\
     &  +\sum_{k=1}^K \sum_{n=1}^N   D_{nn}   E_{kk} w^2_{nk} 
\end{align}
\end{subequations}
\endgroup
It follows that:
\begingroup\small
\begin{subequations}
\label{eqp1w}
\begin{align}
\|X-XWG\|^2_F  & =  Tr(X^TX)-2\sum_{n=1}^N\sum_{k=1}^K C_{kn} w_{nk} \\ & +\sum_{k=1}^K \sum_{n=1}^N   D_{nn}   E_{kk} w^2_{nk} \\
&+\sum_{k=1}^K \sum_{n=1}^N \sum\limits_{\underset{n'\neq n} {n'=1}}^N \sum\limits_{\underset{k'\neq k} {k'=1}}^K  D_{nn'}   E_{k' k} w_{nk} w_{n'k'} 
\end{align}
\end{subequations}
\endgroup
where   $C=GX^TX$, $D=X^TX$ and 
$E=GG^T$.
\\
\\The constraint  $\sum_{k=1}^K\left(1- \sum_{n=1}^N w_{nk}\right)^2$ in  \eqref{eq:CNMFW} can be written as  \eqref{eqp3}.  
\begin{equation}
\label{eqp3w}
    \sum_{k=1}^K\left(1  -2 \sum_{n=1}^N w_{nk} +
\sum_{n=1}^N  \sum\limits_{\underset{n'\neq n} {n'=1}}^N w_{nk}w_{n'k}+\sum_{n=1}^N   w^2_{nk} \right)
\end{equation}

\subsection{Construction of the QUBOs of our problems}

\noindent In what follows we describe how to transform the two problems described in \eqref{eq:CNMFG} and \eqref{eq:CNMFW} to two independent  QUBO problem with respect to $G$ and $W$. In both equations \eqref{eq:CNMFG} and \eqref{eq:CNMFW}, $W$ and $G$ are matrices with real values. Therefore we use the method introduced in \ref{sq:cQUBO} to present these values with binary variables. After that, it is enough to find the two coefficients of each QUBO (linear coefficients $\psi(b)$ and quadratic coefficients $\psi'(b,b')$).

\subsubsection{Binary minimization problem with respect to $G$}

For the problem \eqref{eq:CNMFG} we define the linear coefficients  $\psi(b)$  and the quadratic coefficients $\psi'(b,b')$ as follows: 
\begingroup\small
\begin{subequations}
\label{goalK}
\begin{align}
\psi(b) &= \sum_{n=1}^N  \sum_{k=1}^{K} (- 2( A_{nk}+1) \beta^{k}_{nb}
    +  (B_{kk}+1) (\beta^{k}_{nb})^2 ) \\
\psi'(b,b') &=   2 \sum_{k=1}^{K}  \sum\limits_{\underset{k'\neq k} {k'=1}}^K \sum_{n=1}^N B_{kk'} \beta^{k'}_{nb}\beta^{k}_{nb'} 
    \\ &+ 2 \sum_{k=1}^{K}  \sum_{n=1}^N \sum\limits_{\underset{n'\neq n} {n'=1}}^N\beta^{k}_{nb} \beta^{k}_{n'b'}\\
    & +2\sum_{k=1}^{K} \sum_{n=1}^{N} (B_{kk}+1)    \beta^{k}_{nb} \beta^{k}_{nb'}
\end{align}
\end{subequations}
\endgroup
These formulations follow directly from  \eqref{eqp2} and \eqref{eqp3}.

 \subsubsection{Binary minimization problem with respect to $W$}
 
For the problem \eqref{eq:CNMFW} we define the linear coefficients  $\psi(b)$  and the quadratic coefficients $\psi'(b,b')$ as follows: 
\begingroup\small
\begin{subequations}
\label{eqp1wb}
\begin{align}
\psi(b)   &= \sum_{n=1}^N\sum_{k=1}^K -2(C_{kn}+1) \beta^n_{kb}
       \\ & +\sum_{k=1}^K \sum_{n=1}^N   (D_{nn}   E_{kk}+1) (\beta^n_{kb})^2 \\
\psi'(b,b')  & = \sum_{k=1}^K \sum_{n=1}^N 2(D_{nn}E_{kk} +1) \beta^n_{kb}\beta^n_{kb'}\\
&+ 2\sum_{k=1}^K \sum_{n=1}^N   \sum\limits_{\underset{n'\neq n} {n'=1}}^N (\sum\limits_{\underset{k'\neq k} {k'=1}}^K D_{nn'}E_{k'k} \beta^{n}_{kb}\beta^{n'}_{k'b'} +\beta^{n}_{kb} \beta^{n}_{kb'} )
\end{align}
\end{subequations}
\endgroup
These formulations follow directly from  \eqref{eqp1w} and \eqref{eqp3w}.


\subsection{Embedding in D-WAVE 2000Q }
\begin{figure}[h]
       \centering
       \includegraphics[scale=0.45]{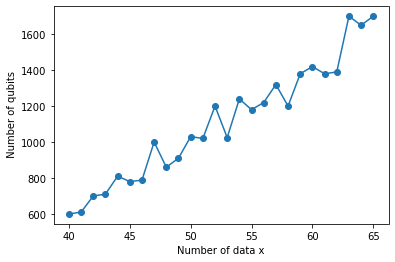}
       \caption{The number of qubits used in D-wave 2000Q after the embedding of a problem completely connected, in the case where each real number is represented by $10$ qubits ($B=9$ and $\psi'(b,b')!=0$ for each $b \neq b'$). }
       \label{fig:num_q}
   \end{figure}

\noindent After having built the two QUBOs of our problems, we notice that the $\psi'(b,b')\neq 0$  coefficient for each pair of qubits $b$ and $b'$. This remark allows us to deduce that the qubits of our problem are completely connected between them.  It means that the chimera graph which represents the connection of the qubits of our problem is completely connected. On the other hand, D-wave 2000Q processes a set of qubits and a set of couplers between some pairs of qubits. The problem here is that there exists qubits in the quantum processor of this computer which are not connected between them.
 So, the execution of our problem directly in D-wave 2000Q is not possible.  To make it feasible we use multiple qubits to represent each variable $b$ of our problems. That means we need to embed our chimera graph into the D-wave 2000Q chimera graph. See the paper \cite{zbinden2020embedding} for more details. On the other hand, to calculate the maximum number of real values can be used in our problems, we consider these two conditions:
\begin{itemize}
    \item According to \ref{sq:cQUBO}, we use $10$ qubits to represent all the real values  $x_{mn} \in [0,1.023]$.
    \item  The chimeras graphs of our problems are completely connected.
\end{itemize}
The figure \ref{fig:num_q} shows the maximum number of real values that we can process using our approach in D-wave 2000Q. Unfortunately, until now, we can't test our approach on a large data set. The matrices that we can factorize using our approach in the D-wave 2000Q quantum computer must not exceed 65 real values.  In other words, if we have a data set $X \in \R^{M \times N}$, it is necessary that $M \times N \leq 65$.

 \section{Results }
 \label{sq:res}
\noindent  In order to test our approach we generate randomly a data set $X \in \R^{M \times N}$ with $M=20$ and $N=3$. In this case we have $N\times M=60 \leq 65$. Therefore we can use our approach in D-wave 2000Q to approximate the two matrices $G$ and $W$. 
Firstly, we build the QUBO of the problem \eqref{eq:CNMFG} and we execute this QUBO in D-Wave 2000Q to find the matrix $G$. Secondly, we use this matrix $G$ to build the QUBO of the problem \eqref{eq:CNMFW} and we execute this QUBO in D-wave 2000Q to find the matrix $W$.  After finding the two matrices,  we represent the two best centers by the red color in figure \ref{result} as  follows:

\begin{figure}[ht]
\centering
\includegraphics[scale=0.25]{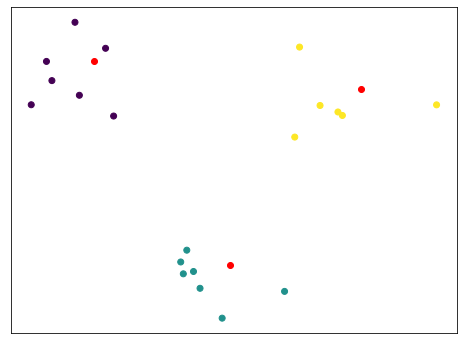}
\includegraphics[scale=0.25]{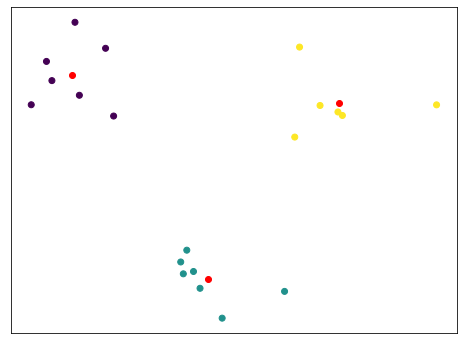}
\caption{Results of the test in D-wave 2000Q, each figures represent one of the best results returned by D-wave 2000Q. The red color represents the centroids and each cluster is represented by a color.}
\label{result}
\end{figure} 
\noindent In figure \ref{result}, we can clearly see that our approach is able to find the right centroids of clusters. These results are very interesting, even if it is on a small data set, because it proves that our approach works well and that the D-wave quantum computer can find the right results that minimize the two  optimization problems \eqref{eq:CNMFG0} and \eqref{eq:CNMFW0} with only one iteration for each problem.  
\\
\\The major limitation of our approach is the number of data that we can handle for each of the two problems deifined in the Equations \eqref{eq:CNMFG0} and \eqref{eq:CNMFW0} respectively. Indeed  the actual quantum computers cannot exceed 65 real values in the data matrix.  This limitation is related to the number of qubit that we can manipulated in the D-wave quantum computer. However, our approach will be a very pertinent solution to solve the problem of computation time if we have quantum computers with a large number of qubits. This limitation will be a distant memory after a period of time. Because with the results of the paper \cite{doiabg9158}, we can move to silicon quantum processors with millions of qubits instead of the current devices with a few qubits. 

\subsection{Run time analysis }
\begin{figure}[h]
       \centering
       \includegraphics[scale=0.3]{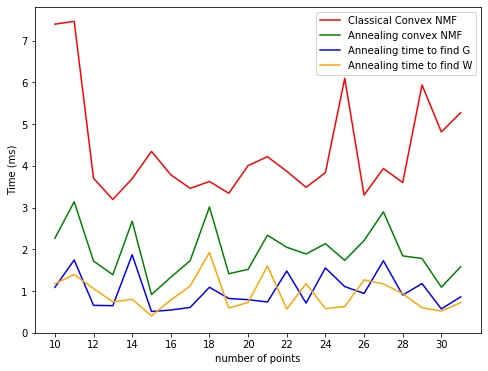}
       \caption{ Total computing  time of Classical Convex-NMF (red color) and Quantum Convex-NMF (green color) as the number of points of the data set. The blue color is the total computing time to find G and the orange to find W. }
       \label{fig:time}
   \end{figure}

\noindent In order to analyze and compare the computing time of our approach with the classical approach, we made several executions on several small randomly generated data sets $(X_1,X_2,\dots, X_{12})$, each data of these data sets  $x\in X_i$ is a vector of  $\R^2$. 
The maximum number of real valuers that we can use in our approach is 65, so we have generated 12 random data sets with different sizes and containing at most 64 reals values. We have run our approach on these data sets and also the classical Convex-NMF approach to compare the computing time of each approach. The results of this analysis are displayed in figure \ref{fig:time} with the red color for classical Convex-NMF  and green for Quantum Convex-NMF  and the blue color presents the time to find the matrix $G$ and orange to find $W$ in D-wave 2000Q.   
   
\noindent The green curve in the figure is simply the sum of the two curves orange and blue. This is intrinsically due to our alternative strategy, in order to find the results using our approach, we solve the first QUBO to find $G$ and then the last QUBO to find $W$. Therefore, the total running time on D-wave 2000Q is simply calculated as the sum of the running times of these two problems. By comparing the total time spent to solve the problem with our approach (the green curve) and the time taken by the classical convex-NMF approach (the red curve), we can see clearly that our approach is very fast than the classical approach.

 \section{Conclusion}
 
\noindent For the last few years, the power of the adiabatic quantum computers has become a relevant solution for highly complex algorithms such as machine learning ones. In this paper, we have proposed a new approach for the Convex-NMF algorithm in D-wave 2000Q. In this approach we have proposed to decompose the Convex-NMF optimization problem on two QUBOs problems: the first to find the matrix $G$ and the second to find the matrix $W$ which minimizes the norm difference between the data matrix $X$ and the matrix product $XWG$  where all elements of $G$ and $W$ are non-negative and on the rows/columns they sum up to 1. To work with real values  we used a transformation  proposed in \cite{ottaviani2018low}, this transformation allows to find a binary representation of the problem from a problem with real variables. Before testing our approach on D-wave 2000Q, we made a study to find the maximum number of real data to use in our problems, this study is based on the number of qubits of D-wave 2000Q, the connection of those qubits between them (chimera graph architecture) and according to the number of qubits used to represent each real variable in the QUBO. Our approach is tested on a small data set generated randomly, to demonstrate that our approach works well and that the D-wave quantum computer is capable to find the right clusters of the data set. Also we have made several tests to demonstrate that our quantum approach is faster than the classical  method. Until today, we cannot test our approach on large data sets, because the number of qubits in D-wave quantum computers is limited. We hope that in the coming years the number of qubits in quantum computers will increase.

\medskip

\bibliographystyle{plain}
\bibliography{bib}

\end{document}